\pdfoutput=1

\documentclass[11pt]{article}

\usepackage{acl}

\usepackage{times}
\usepackage{latexsym}

\usepackage[T1]{fontenc}

\usepackage[utf8]{inputenc}

\usepackage{microtype}
\usepackage{graphicx}
\usepackage{amsmath,amssymb,bm}
\usepackage{xspace}
\usepackage{tabularx}
\usepackage{booktabs}
\usepackage{subcaption}
\usepackage{colortbl}
\usepackage{makecell}
\usepackage{float}
\usepackage{enumitem}

\newcommand{\model}{\textsc{TableFormer}\xspace}
\definecolor{dg}{rgb}{0.0, 0.5, 0.0}
\title{\textsc{TableFormer}: Robust Transformer Modeling for Table-Text Encoding}

\author{Jingfeng Yang$^{\thanks{  ~~Work done during an internship at Google.}}$ \quad Aditya Gupta$^{\dagger}$ \quad  Shyam Upadhyay$^{\dagger}$ \\ {\bf Luheng He}$^{\dagger}$ \quad {\bf Rahul Goel}$^{\dagger}$ \quad {\bf Shachi Paul} $^{\dagger}$\\
  $^{\star}$Georgia Institute of Technology\\
  $^\dagger$Google Assistant\\
  {\tt jingfengyangpku@gmail.com } \\
  {\tt tableformer@google.com}
  }

\begin{document}
\maketitle
\begin{abstract}
Understanding tables is an important aspect of natural language understanding. Existing models for table understanding require linearization of the table structure, where row or column order is encoded as an \emph{unwanted} bias. Such spurious biases make the model vulnerable to row and column order perturbations. Additionally, prior work has not thoroughly modeled the table structures or table-text alignments, hindering the table-text understanding ability. In this work, we propose a robust and structurally aware table-text encoding architecture \textsc{TableFormer}, where tabular structural biases are incorporated completely through learnable attention biases. \textsc{TableFormer} is (1) strictly invariant to row and column orders, and, (2) could understand tables better due to its tabular inductive biases.  Our evaluations showed that \textsc{TableFormer} outperforms strong baselines in all settings on SQA, WTQ and \textsc{TabFact} table reasoning datasets, and achieves state-of-the-art performance on SQA, especially when facing answer-invariant row and column order perturbations (6\% improvement over the best baseline), because previous SOTA models' performance drops by 4\% - 6\% when facing such perturbations while \textsc{TableFormer} is not affected.\footnote{Code has been released at \url{https://github.com/google-research/tapas/blob/master/TABLEFORMER.md}}
\end{abstract}

\section{Introduction}

Recently, semi-structured data (e.g. variable length tables without a fixed data schema) has attracted more attention because of its ubiquitous presence on the web. On a wide range of various table reasoning tasks, Transformer based architecture along with pretraining has shown to perform well \cite{ eisenschlos2021mate, liu2021tapex}. 

\begin{figure}[t!]
\centering
\footnotesize
\begin{subfigure}{\linewidth}
\begin{table}[H]
\centering
\footnotesize
\begin{tabular}{|c|c|}
\hline
\rowcolor[HTML]{FFCCC9} 
Title                     & Length        \\
\hline 
\rowcolor[HTML]{DAE8FC} 
\textbf{Screwed Up}  & \textbf{5:02} \\
\rowcolor[HTML]{DAE8FC} 
Ghetto Queen & 5:00  \\
\hline
\end{tabular}
\end{table}
\vspace{-1em}
\textbf{Question:} Of all song lengths, which one is the longest? \\
\textbf{Gold Answer:} 5:02 \\
\textbf{TAPAS:} \textcolor{red}{5:00} \\
\textbf{TAPAS after row order perturbation:} \textcolor{dg}{5:02} \\
\textbf{\textsc{TableFormer}:} \textcolor{dg}{5:02}
\caption{\textsc{Tapas} predicts incorrect answer based on the original table, while it gives the correct answer if the first row is moved to the end of the table.}
\end{subfigure}

\begin{subfigure}{\linewidth} 
\begin{table}[H]
\centering
\footnotesize
\begin{tabular}{|c|c|c|c|}
\hline
\rowcolor[HTML]{FFCCC9} 
Nation           & Gold       & Silver     & Bronze     \\ \hline
\rowcolor[HTML]{DAE8FC} 
Great Britain    & 2          & 1          & 2          \\ \hline
\rowcolor[HTML]{DAE8FC} 
\textbf{Spain}   & \textbf{1} & \textbf{2} & \textbf{0} \\ \hline
\rowcolor[HTML]{DAE8FC} 
\textbf{Ukraine} & \textbf{0} & \textbf{2} & \textbf{0} \\ \hline
\end{tabular}
\end{table}
\vspace{-1em}
\textbf{Question:} Which nation received 2 silver medals? \\
\textbf{Gold Answer:} Spain, Ukraine\\
\textbf{TAPAS:} \textcolor{red}{Spain} \\
\textbf{\textsc{TableFormer}:} \textcolor{dg}{Spain, Ukraine} \\
\textbf{\textsc{TableFormer} w/o a proposed structural bias:} \textcolor{red}{Spain}
\caption{\textsc{Tapas} gives incomplete answer due to its limited cell grounding ability.}
\end{subfigure}
\caption{\label{fig:example} Examples showing the limitations of existing models (a) vulnerable to perturbations, and (b) lacking structural biases. In contrast, our proposed \textsc{TableFormer} predicts correct answers for both questions.}
\vspace{-2em}
\end{figure}

In a nutshell, prior work used the Transformer architecture in a BERT like fashion by serializing tables or rows into word sequences \cite{yu2020grappa, liu2021tapex}, where original position ids are used as positional information. Due to the usage of row/column ids and global position ids, prior strategies to linearize table structures introduced spurious row and column order biases \cite{herzig-etal-2020-tapas, eisenschlos-etal-2020-understanding, eisenschlos2021mate, zhang2020table, yin20acl}. Therefore, those models are vulnerable to row or column order perturbations.
But, ideally, the model should make consistent predictions regardless of the row or column ordering for all practical purposes. For instance, in Figure \ref{fig:example}, the predicted answer of \textsc{Tapas} model \cite{herzig-etal-2020-tapas} for Question (a) \emph{``Of all song lengths, which one is the longest?''} based on the original table is \emph{``5:00''}, which is incorrect. However, if the first row is adjusted to the end of the table during inference, the model gives the correct length \emph{``5:02''} as answer. This probing example shows that the model being aware of row order information is inclined to select length values to the end of the table due to spurious training data bias. In our experiments on the SQA dataset, \textsc{Tapas} models exhibit a 4\% - 6\% (Section \ref{perturb-sec}) absolute performance drop when facing such answer-invariant perturbations.

Besides, most prior work \cite{chen2019tabfact, yin20acl} did not incorporate enough structural biases to models to address the limitation of sequential Transformer architecture, while others inductive biases which are either too strict \cite{zhang2020table, eisenschlos2021mate} or computationally expensive \cite{yin20acl}.


To this end, we propose \textsc{TableFormer}, a Transformer architecture that is robust to row and column order perturbations, by incorporating structural biases more naturally. \textsc{TableFormer} relies on 13 types of task-independent table$\leftrightarrow$text attention biases that respect the table structure and table-text relations.
For Question (a) in Figure~\ref{fig:example}, \textsc{TableFormer} could predict the correct answer regardless of perturbation, because the model could identify the same row information with our \emph{``same row''} bias,  avoiding spurious biases introduced by row and global positional embeddings. 
For Question (b), \textsc{Tapas} predicted only partially correct answer, while \textsc{TableFormer} could correctly predict \emph{``Spain, Ukraine''} as answers. That's because our \emph{``cell to sentence''} bias could help table cells ground to the paired sentence. Detailed attention bias types are discussed in Section \ref{perturb-sec}.

Experiments on 3 table reasoning datasets show that \textsc{TableFormer} consistently outperforms original \textsc{Tapas} in all pretraining and intermediate pretraining settings with fewer parameters. Also, \textsc{TableFormer}'s invariance to row and column perturbations, leads to even larger improvement over those strong baselines when tested on perturbations. Our contributions are as follows:
\begin{itemize}
    \item We identified the limitation of current table-text encoding models when facing row or column perturbation. 
    \item We propose \textsc{TableFormer}, which is guaranteed to be  invariant to row and column order perturbations, unlike current models.  
    \item \textsc{TableFormer} encodes table-text structures better, leading to SoTA performance on SQA dataset, and ablation studies show the effectiveness of the introduced inductive biases.
\end{itemize}

\section{Preliminaries: \textsc{Tapas} for Table Encoding}
In this section, we discuss \textsc{TAPAS} which serves as the backbone of the recent state-of-the-art table-text encoding architectures.
\textsc{Tapas} \cite{herzig-etal-2020-tapas} uses Transformer architecture in a BERT like fashion to pretrain and finetune on tabular data for table-text understanding tasks. This is achieved by using linearized table and texts for masked language model pre-training. In the fine-tuning stage, texts in the linearized table and text pairs are queries or statements in table QA or table-text entailment tasks, respectively. 

Specifically, \textsc{Tapas} uses the tokenized and flattened text and table as input, separated by \texttt{[SEP]} token, and prefixed by \texttt{[CLS]}. Besides token, segment, and global positional embedding introduced in BERT \cite{devlin2018bert}, it also uses rank embedding for better numerical understanding. Moreover, it uses column and row embedding to encode table structures. 

Concretely, for any table-text linearized sequence $S = \{v_1, v_2, \cdots, v_n\}$, where $n$ is the length of table-text sequence, the input to \textsc{Tapas} is summation of embedding of the following:
\begin{equation*} 
\begin{split}
\text{token ids } (W) & = \{w_{v_1}, w_{v_2}, \cdots, w_{v_n}\} \\
\text{positional ids } (B) & = \{b_1, b_2, \cdots, b_n\}\\ 
\text{segment ids } (G) & = \{g_{seg_1}, g_{seg_2}, \cdots, g_{seg_n}\} \\ 
\text{column ids } (C) & = \{c_{col_1}, c_{col_2}, \cdots, c_{col_n}\}\\ 
\text{row ids } (R) & = \{r_{row_1}, r_{row_2}, \cdots, r_{row_n}\}  \\ 
\text{rank ids } (Z) & = \{z_{rank_1}, z_{rank_2}, \cdots, z_{rank_n}\}
\end{split}
\end{equation*}

where $seg_i,~col_i,~row_i,~rank_i$ correspond to the segment, column, row, and rank id for the $i$th token, respectively.

\begin{figure*}
\begin{center}
\includegraphics[width=1\linewidth]{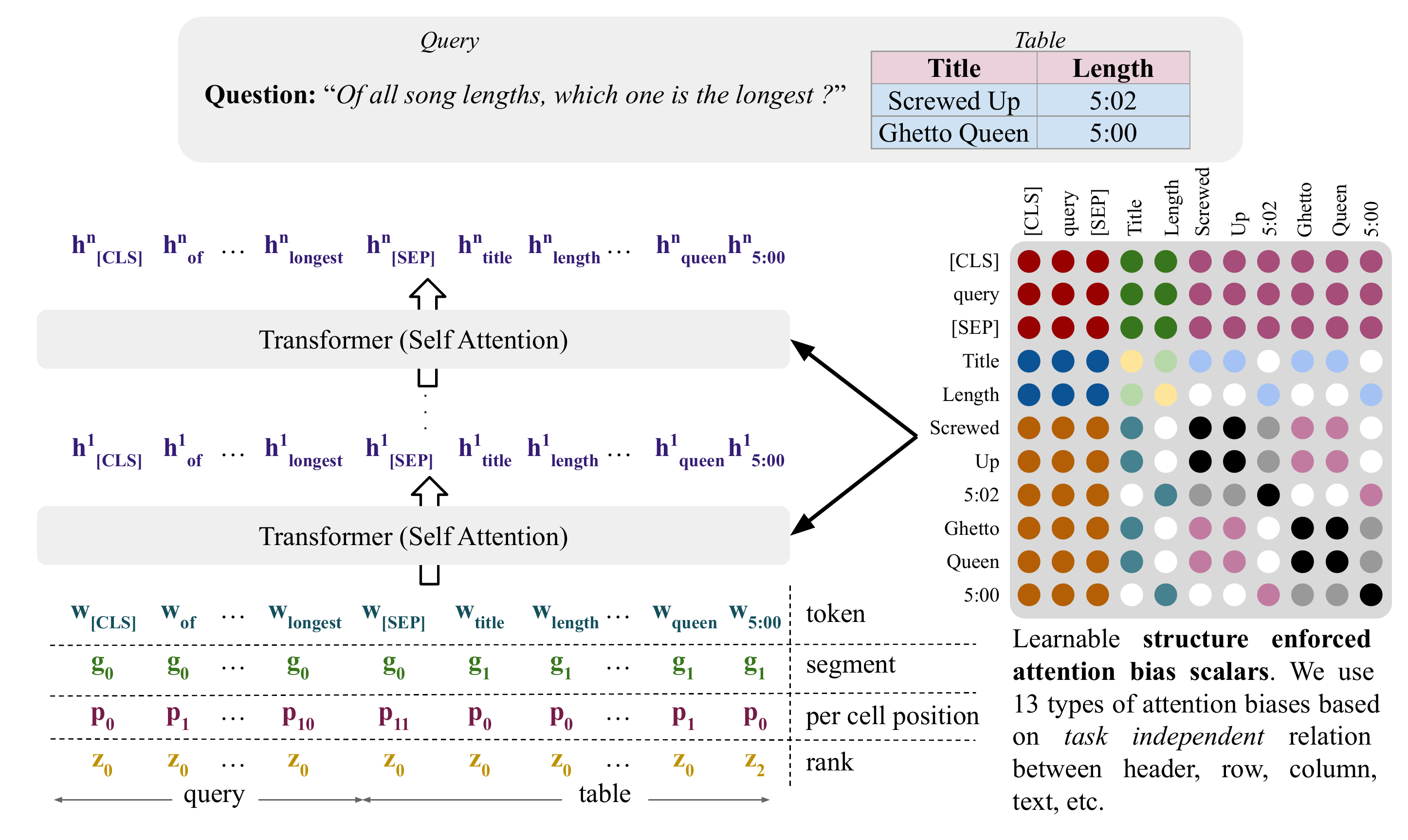}
\end{center}
\caption{\label{fig:model}  \textsc{TableFormer} input and attention biases in the self attention module. This example corresponds to table (a) in Figure \ref{fig:example} and its paired question ``query''. Different colors in the attention bias matrix denote different types of task independent biases derived based on the table structure and the associated text.
}

\end{figure*}

As for the model, \textsc{Tapas} uses  BERT's self-attention architecture \cite{vaswani2017attention} off-the-shelf. Each Transformer layer includes a multi-head self-attention sub-layer, where each token attends to all the tokens.  Let the layer input $H = [h_1, h_2, \cdots, h_n]^\top \in \mathbb{R}^{n\times d}$ corresponding to $S$, where $d$ is the hidden dimension, and $h_i\in \mathbb{R}^{d\times 1}$ is the hidden representation at position $i$. For a single-head self-attention sub-layer, the input $H$ is projected by three matrices $W^Q \in \mathbb{R}^{d\times d_K}$, $W^K \in \mathbb{R}^{d\times d_K}$, and $W^V \in \mathbb{R}^{d\times d_V}$ to the corresponding representations $Q$, $K$, and $V$: 
\begin{equation}
\label{eqp}
\begin{aligned}
Q = H W^Q, \quad V = H W^V, \quad K = H W^K
\end{aligned}
\end{equation}

Then, the output of this single-head self-attention sub-layer is calculated as: 
\begin{equation}
\label{eqatt}
\begin{aligned}
\text{Attn}(H) = \text{softmax} (\frac{QK^\top}{\sqrt{d_K}})V
\end{aligned}
\end{equation}

\section{\textsc{TableFormer}: Robust Structural Table Encoding}

As shown in Figure~\ref{fig:model}, \textsc{TableFormer} encodes the general table structure along with the associated text by introducing task-independent relative attention biases for table-text encoding to facilitate the following:  (a) structural inductive bias for better table understanding and table-text alignment, (b) robustness to table row/column perturbation. 

\paragraph{Input of \textsc{TableFormer}.}
\label{sec:input}

\textsc{TableFormer} uses the same token embeddings $W$, segment embeddings $G$, and rank embeddings $Z$ as \textsc{Tapas}. However, we make 2 major modifications:
\paragraph{1) No row or column ids.} We do not use row embeddings $R$ or column embeddings $C$ to avoid any potential spurious row and column order biases.
\paragraph{2) Per cell positional ids.} To further remove any inter-cell order information, global positional embeddings $B$ are replaced by per cell positional embeddings $P = \{p_{pos_1}, p_{pos_2}, \cdots, p_{pos_n}\}$, where we follow \citet{eisenschlos2021mate} to reset the index of positional embeddings at the beginning of each cell, and $pos_i$ correspond to the per cell positional id for the $i$th token.

\paragraph{Positional Encoding in \textsc{TableFormer}.}

Note that the Transformer model either needs to specify different positions in the input (i.e. absolute positional encoding of \citet{vaswani2017attention}) or encode the positional dependency in the layers (i.e. relative positional encoding of \citet{shaw2018self}). 

\model does not consume any sort of column and row order information in the input. The main intuition is that, for cells in the table, the only useful positional information is whether two cells are in the same row or column and the column header of each cell, instead of the absolute order of the row and column containing them. Thus, inspired by relative positional encoding \cite{shaw2018self} and graph encoding \cite{ying2021transformers}, we capture this with a  same column/row relation as one kind of relative position between two linearized tokens. Similarly, we uses 12 such table-text structure relevant relations (including same cell, cell to header and so on) and one extra type representing all other relations not explicitly defined. All of them are introduced in the form of learnable attention bias scalars. 

Formally, we consider a function $\phi (v_i, v_j): V \times V \rightarrow \mathbb{N}$, which measures the relation between $v_i$ and $v_j$ in the sequence ($v_i,v_j\in S$). The function $\phi$ can be defined by any relations between the tokens in the table-text pair. 


\paragraph{Attention Biases in \textsc{TableFormer}.}
\label{bias-sec}
In our work, $\phi (v_i, v_j)$ is chosen from 13 bias types, corresponding to 13 table-text structural biases. The attention biases are applicable to any table-text pair and can be used for any downstream task:
\begin{itemize}
    \item \emph{``same row''} identifies the same row information without ordered row id embedding or global positional embedding, which help the model to be invariant to row perturbations,
    \item \emph{``same column''}, \emph{``header to column cell''}, and \emph{``cell to column header''} incorporates the same column information without ordered column id embedding,
    \item \emph{``cell to column header''} makes each cell aware of its column header without repeated column header as features,
    \item  \emph{``header to sentence''} and \emph{``cell to sentence''} help column grounding and cell grounding of the paired text,
    \item \emph{``sentence to header''}, \emph{``sentence to cell''}, and \emph{``sentence to sentence''} helps to understand the sentence with the table as context,
    \item  \emph{``header to same header''} and \emph{``header to other header''} for better understanding of table schema, and \emph{``same cell bias''} for cell content understanding.
\end{itemize}
 Note that, each cell can still attend to other cells in the different columns or rows through \emph{``others''} instead of masking them out strictly. 

We assign each bias type a learnable scalar, which will serve as a bias term in the self-attention module. Specifically, each self-attention head in each layer have a set of learnable scalars $\{b_1, b_2, \cdots, b_{13}\}$ corresponding to all types of introduced biases. For one head in one self-attention sub-layer of \textsc{TableFormer}, Equation \ref{eqatt} in the Transformer is replaced by:

\begin{equation}
\label{eqtatt}
\begin{aligned}
\bar{A} = \frac{QK^\top}{\sqrt{d_K}},  \quad A = \bar{A} + \hat{A}
\end{aligned}
\end{equation}
\begin{equation}
\begin{aligned}
\text{Attn}(H) = \text{softmax} (A)V
\end{aligned}
\end{equation}

where $\bar{A}$ is a matrix capturing the similarity between queries and keys, $\hat{A}$ is the Attention Bias Matrix, and $\hat{A}_{i,j} = b_{\phi(v_i, v_j)}$. 

\paragraph{Relation between \textsc{TableFormer} and ETC.}
\label{para:ETC}
ETC~\cite{ainslie2020etc} uses vectors to represent relative position labels, although not directly applied to table-text pairs due to its large computational overhead \cite{eisenschlos2021mate}. \textsc{TableFormer} differs from ETC in the following aspects (1) ETC uses relative positional embeddings while \textsc{TableFormer} uses attention bias scalars. In practice, we observed that using relative positional embeddings increases training time by more than 7x, (2) ETC uses global memory and local attention, while \textsc{TableFormer} uses pairwise attention without any global memory overhead, (3) ETC uses local sparse attention with masking, limiting its ability to attend to all tokens, (4) ETC did not explore table-text attention bias types exhaustively. Another table encoding model \textsc{MATE}~\cite{eisenschlos2021mate} is vulnerable to row and column perturbations, and  shares limitation (3) and (4).


\section{Experimental Setup}

\subsection{Datasets and Evaluation}
We use the following datasets in our experiments.
\paragraph{Table Question Answering.} For the table QA task, we conducted experiments on WikiTableQuestions (\textsc{WTQ}) \cite{pasupat2015compositional} and Sequential QA (SQA) \cite{iyyer2017search} datasets. \textsc{WTQ} was crowd-sourced based on complex questions on Wikipedia tables. SQA is composed of $6,066$ question sequences
(2.9 question per sequence on average), constructed by decomposing a subset of highly compositional \textsc{WTQ} questions. 
\paragraph{Table-Text Entailment.} For the table-text entailment task, we used \textsc{TabFact} dataset \cite{chen2019tabfact}, where the tables were extracted from Wikipedia and the sentences were written by crowd workers.  Among total 118, 000 sentences, each one is a positive (entailed) or negative sentence.

\paragraph{Perturbation Evaluation Set.} For SQA and \textsc{TabFact}, we also created new test sets to measure models' robustness to answer-invariant row and column perturbations during inference. Specifically, row and column orders are randomly perturbed for all tables in the standard test sets.\footnote{We fixed perturbation random seeds to make our results reproducible.}

\paragraph{Pre-training} 
All the models are first tuned on the \emph{Wikipidia} text-table pretraining dataset \cite{herzig-etal-2020-tapas}, optionally tuned on synthetic dataset at an intermediate stage (``inter'') \cite{eisenschlos-etal-2020-understanding}, and finally fine-tuned on the target dataset.  To get better performance on \textsc{WTQ}, we follow \citet{herzig-etal-2020-tapas} to further pretrain on SQA dataset after the intermediate pretraining stage in the ``inter-sqa'' setting.

\paragraph{Evaluation}
For SQA, we report the cell selection accuracy for all questions (\textsc{ALL}) using the official evaluation script, cell selection accuracy for all sequences (\textsc{SEQ}), and the denotation accuracy for all questions ($\text{ALL}_{\text{d}}$). To evaluate the models' robustness in the instance level after perturbations, we also report a lower bound of example prediction variation percentage: 
\begin{equation}
VP= \frac{(\text{t2f} + \text{f2t})}{(\text{t2t}+\text{t2f}+\text{f2t}+\text{f2f})}    
\end{equation}
where t2t, t2f, f2t, and f2f represents how many example predictions turning from correct to correct, from correct to incorrect, from incorrect to correct and from incorrect to incorrect, respectively, after perturbation.
We report denotation accuracy on \textsc{WTQ} and binary classification accuracy on \textsc{TabFact} respectively.
 
\subsection{Baselines} We use $\textsc{Tapas}_{\text{BASE}}$ and $\textsc{Tapas}_{\text{LARGE}}$ as baselines, where Transformer architectures are exactly same as $\textsc{BERT}_{\text{BASE}}$ and $\textsc{BERT}_{\text{LARGE}}$ \cite{devlin2018bert}, and parameters are initialized from $\textsc{BERT}_{\text{BASE}}$ and $\textsc{BERT}_{\text{LARGE}}$ respectively. Correspondingly, we have our $\textsc{TableFormer}_{\text{BASE}}$ and $\textsc{TableFormer}_{\text{LARGE}}$, where attention bias scalars are initialized to zero, and all other parameters are initialized from $\textsc{BERT}_{\text{BASE}}$ and $\textsc{BERT}_{\text{LARGE}}$.

\subsection{Perturbing Tables as Augmented Data}
\label{sec:aug}
\emph{Could we alleviate the spurious ordering biases by data augmentation alone, without making any modeling changes?} To answer this, we train another set of models by augmenting the training data for TAPAS through random row and column order perturbations.\footnote{By perturbation, we mean shuffling row and columns instead of changing/swapping content blindly.} 

For each table in the training set, we randomly shuffle all rows and columns (including corresponding column headers), creating a new table with the same content but different orders of rows and columns. Multiple perturbed versions of the same table were created by repeating this process $\{1, 2, 4, 8, 16\}$ times with different random seeds. For table QA tasks, selected cell positions are also adjusted as final answers according to the perturbed table. The perturbed table-text pairs are then used to augment the data used to train the model. During training, the model takes data created by one specific random seed in one epoch in a cyclic manner. 

\begin{table}
\setlength\tabcolsep{4pt}
\scriptsize
\centering
\begin{tabular}{lccc|cc}
\toprule
& \multicolumn{3}{c|}{\bf Before Perturb} & \multicolumn{2}{c}{\bf After Perturb}   \\\midrule
                                             & ALL           & SEQ           & $\text{ALL}_{\text{d}}$ & ALL           & $VP$                                                \\\midrule
\citet{herzig-etal-2020-tapas}                         & 67.2          & 40.4          &       --           &         --      &                                 --                                          \\
\citet{eisenschlos-etal-2020-understanding}                   & 71.0          & 44.8          &           --       &         --      &           --                                                              \\
\citet{eisenschlos2021mate} & 71.7 & 46.1 & -- & -- & --\\
\citet{liu2021tapex}                           &       --        &     --          & 74.5             &      --         &          --                                                                           \\\midrule

$\textsc{Tapas}_{\text{BASE}}$                                  & 61.1          & 31.3          &         --         & 57.4          & 14.0\% \\
$\textsc{TableFormer}_{\text{BASE}}$                             & \textbf{66.7} & \textbf{39.7} &          --        & \textbf{66.7} & \textbf{0.2\%}                                                \\\midrule

$\textsc{Tapas}_{\text{LARGE}}$                                 & 66.8          & 39.9          &         --         & 60.5          & 15.1\%                                                 \\
$\textsc{TableFormer}_{\text{LARGE}}$                           & \textbf{70.3} & \textbf{44.8} &          --        & \textbf{70.3} & \textbf{0.1\%}                                                 \\\midrule

$\textsc{Tapas}_{\text{BASE}}$   inter             & 67.5          & 38.8          &      --            & 61.0          & 14.3\%                                                \\
$\textsc{TableFormer}_{\text{BASE}}$ inter  & \textbf{69.4} & \textbf{43.5} &  --                & \textbf{69.3} & \textbf{0.1\%}                                                \\\midrule

$\textsc{Tapas}_{\text{LARGE}}$  inter      & 70.6          & 43.9          &      --            & 66.1          & 10.8\%\\
$\textsc{TableFormer}_{\text{LARGE}}$ inter & \textbf{72.4} & \textbf{47.5} & \textbf{75.9}    & \textbf{72.3} & \textbf{0.1\%}      
\\\bottomrule
\end{tabular}
\caption{\label{tab:widgets1} Results on SQA test set before and after perturbation during inference (median of 5 runs). ALL is cell selection accuracy, SEQ is cell selection accuracy for all question sequences, $\text{ALL}_{\text{d}}$ is denotation accuracy for all questions (reported to compare with \citet{liu2021tapex}). $VP$ is model prediction variation percentage after perturbation. Missing values are those not reported in the original paper. }
\end{table}

\section{Experiments and Results}

\begin{table*}
\footnotesize
\centering
\begin{tabular}{cccccc|cccc}
\toprule
& \multicolumn{5}{c|}{ \bf Before Perturb} & \multicolumn{4}{c}{\bf After Perturb} \\\midrule
& dev           & test          & $\text{test}_{\text{simple}}$  & $\text{test}_{\text{complex}}$ & $\text{test}_{\text{small}}$   & test       & $\text{test}_{\text{simple}}$  & $\text{test}_{\text{complex}}$       & $\text{test}_{\text{small}}$   \\\midrule
\citet{eisenschlos-etal-2020-understanding} & 81.0  & 81.0  & 92.3  & 75.6 & 83.9  & --  &  --  &  -- &  -- \\
\citet{eisenschlos2021mate} & -- & 81.4 & -- & -- & -- & -- & -- & -- & -- \\\midrule
$\textsc{Tapas}_{\text{BASE}}$                                           & 72.8          & 72.3          & 84.8          & 66.2          & 74.4 & 71.2 & 83.4 & 65.2 & 72.5          \\
$\textsc{TableFormer}_{\text{BASE}}$                                   & \textbf{75.1} & \textbf{75.0} & \textbf{88.2} & \textbf{68.5} & \textbf{77.1}                                       & \textbf{75.0}                                       & \textbf{88.2}                                       & \textbf{68.5}                                       & \textbf{77.1} \\\midrule
$\textsc{Tapas}_{\text{LARGE}}$                                         & 74.7          & 74.5          & 86.6          & 68.6          & 76.8                                                & 73.7                                                & 86.0                                                & 67.7                                                & 76.1          \\
$\textsc{TableFormer}_{\text{LARGE}}$                                 & \textbf{77.2} & \textbf{77.0} & \textbf{90.2} & \textbf{70.5} & \textbf{80.3}                                       & \textbf{77.0}                                       & \textbf{90.2}                                       & \textbf{70.5}                                       & \textbf{80.3} \\\midrule
$\textsc{Tapas}_{\text{BASE}}$ inter                           & 78.4          & 77.9          & 90.1          & 71.9          & 80.5                                                & 76.8                                                & 89.5                                                & 70.5                                                & 79.7          \\
$\textsc{TableFormer}_{\text{BASE}}$ inter                     & \textbf{79.7} & \textbf{79.2} & \textbf{91.6} & \textbf{73.1} & \textbf{81.7}                                       & \textbf{79.2}                                       & \textbf{91.6}                                       & \textbf{73.1}                                       & \textbf{81.7} \\\midrule
$\textsc{Tapas}_{\text{LARGE}}$  inter                          & 80.6          & 80.6          & 92.0          & 74.9          & 83.1                                                & 79.2                                                & 91.7                                                & 73.0                                                & 83.0          \\
$\textsc{TableFormer}_{\text{LARGE}}$ inter                    & \textbf{82.0} & \textbf{81.6} & \textbf{93.3} & \textbf{75.9} & \textbf{84.6}                                       & \textbf{81.6}                                       & \textbf{93.3}                                       & \textbf{75.9}                                       & \textbf{84.6}
\\\bottomrule
\end{tabular}
\caption{\label{tab:widgets2}Binary classification accuracy on \textsc{TabFact} development and 4 splits of test set, as well as performance on test sets with our perturbation evaluation. Median of 5 independent runs are reported. Missing values are those not reported in the original paper.}
\end{table*}

\begin{table}
\footnotesize
\centering
\begin{tabular}{ccc}
\toprule
  Model                                           & dev           & test          \\\midrule
\citet{herzig-etal-2020-tapas}                              & --             & 48.8          \\
\citet{eisenschlos2021mate} & --             & 51.5          \\\midrule
$\textsc{Tapas}_{\text{BASE}}$                                       & 23.6          & 24.1          \\
$\textsc{TableFormer}_{\text{BASE}}$                                  & \textbf{34.4} & \textbf{34.8} \\\midrule
$\textsc{Tapas}_{\text{LARGE}}$                                        & 40.8          & 41.7          \\
$\textsc{TableFormer}_{\text{LARGE}}$                                 & \textbf{42.5} & \textbf{43.9} \\\midrule
$\textsc{Tapas}_{\text{BASE}}$  inter-sqa                       & 44.8          & 45.1          \\
$\textsc{TableFormer}_{\text{BASE}}$ inter-sqa                 & \textbf{46.7} & \textbf{46.5} \\\midrule
$\textsc{Tapas}_{\text{LARGE}}$  inter-sqa                      & 49.9          & 50.4          \\
$\textsc{TableFormer}_{\text{LARGE}}$ inter-sqa                & \textbf{51.3} & \textbf{52.6}
\\\bottomrule
\end{tabular}
\caption{\label{tab:widgets3} Denotation accuracy on \textsc{WTQ} development and test set. Median of 5 independent runs are reported.}
\vspace{-2em}
\end{table}

Besides standard testing results to compare \textsc{TableFormer} and baselines, we also answer the following questions through experiments:
\begin{itemize}
    \item How robust are existing (near) state-of-the-art table-text encoding models to semantic preserving perturbations in the input?
    \item How does \textsc{TableFormer} compare with existing table-text encoding models when tested on similar perturbations, both in terms of performance and robustness?
    \item Can we use perturbation based data augmentation to achieve robustness at test time?
    \item Which attention biases in \textsc{TableFormer} contribute the most to performance?
\end{itemize}

\subsection{Main Results}
Table \ref{tab:widgets1}, \ref{tab:widgets2}, and \ref{tab:widgets3} shows \textsc{TableFormer} performance on \textsc{SQA, TabFact}, and \textsc{WTQ}, respectively. As can be seen, \model outperforms corresponding \textsc{Tapas} baseline models in all settings on SQA and \textsc{WTQ} datasets, which shows the general effectiveness of \textsc{TableFormer}'s structural biases in Table QA datasets. Specifically, $\textsc{TableFormer}_{\text{LARGE}}$ combined with intermediate pretraining achieves new state-of-the-art performance on SQA dataset.

Similarly, Table \ref{tab:widgets2} shows that \textsc{TableFormer} also outperforms \textsc{Tapas} baseline models in all settings, which shows the effectiveness of \textsc{TableFormer} in the table entailment task.
Note that, \citet{liu2021tapex} is not comparable to our results, because they used different pretraining data, different pretraining objectives, and \textsc{BART NLG} model instead of \textsc{BERT NLU} model. But \textsc{TableFormer} attention bias is compatible with BART model. 

\subsection{Perturbation Results}
\label{perturb-sec}
One of our major contributions is to systematically evaluate models' performance when facing row and column order perturbation in the testing stage. 

Ideally, model predictions should be consistent on table QA and entailment tasks when facing such perturbation, because the table semantics remains the same after perturbation.

However, in Table~\ref{tab:widgets1} and \ref{tab:widgets2}, we can see that in our perturbed test set, performance of all \textsc{Tapas} models drops significantly in both tasks. \textsc{Tapas} models drops by at least 3.7\% and up to 6.5\% in all settings on SQA dataset in terms of ALL accuracy, while our \textsc{TableFormer} being strictly invariant to such row and column order perturbation leads to no drop in performance.\footnote{In SQA dataset, there is at most absolute 0.1\% performance drop because of some bad data point issues. Specifically, some columns in certain tables are exactly the same, but the ground-truth selected cells are in only one of such columns. \textsc{TableFormer} would select from one column randomly. } Thus, in the perturbation setting, \textsc{TableFormer} outperforms all \textsc{Tapas} baselines even more significantly, with at least 6.2\% and 2.4\% improvements on SQA and \textsc{TabFact} dataset, respectively. In the instance level, we can see that, with \textsc{Tapas}, there are many example predictions changed due to high $VP$, while there is nearly no example predictions changed with \textsc{TableFormer} (around zero $VP$). 

\subsection{Model Size Comparison}
We compare the model sizes of \textsc{TableFormer} and \textsc{Tapas} in Table \ref{tab:widgets4}. We added only a few attention bias scalar parameters (13 parameters per head per layer) in \textsc{TableFormer}, which is negligible compared with the BERT model size. Meanwhile, we delete two large embedding metrics (512 row ids and 512 column ids). Thus, \textsc{TableFormer} outperforms \textsc{Tapas} with fewer parameters.

\begin{table}[t]
\small
\centering
\begin{tabular}{cc}
\toprule
Model & Number of parameters            \\\midrule
$\textsc{Tapas}_{\text{BASE}}$                                    & 110 M                                                    \\\hline
$\textsc{TableFormer}_{\text{BASE}}$                             & \makecell{110 M - 2*512*768 \\ + 12*12*13 = \\110 M - 0.8 M + 0.002 M}   \\\midrule
$\textsc{Tapas}_{\text{LARGE}}$                                   & 340 M                                                   \\\hline
$\textsc{TableFormer}_{\text{LARGE}}$                             & \makecell{340 M - 2*512*1024 \\+ 24*16*13 = \\340 M - 1.0 M + 0.005M} 
\\\bottomrule
\end{tabular}
\caption{\label{tab:widgets4}Model size comparison.}
\end{table}

\subsection{Analysis of \textsc{TableFormer} Submodules}
In this section, we experiment with several variants of \textsc{TableFormer} to understand the effectiveness of its submodules. The performance of all variants of \textsc{Tapas} and \textsc{TableFormer} that we tried on the SQA development set is shown in Table \ref{tab:widgets5}.

\paragraph{Learnable Attention Biases v/s Masking.} Instead of adding learnable bias scalars, we mask out some attention scores to restrict attention to those tokens in the same columns and rows, as well as the paired sentence, similar to \citet{zhang2020table} (SAT).  We can see that $\textsc{Tapas}_{\text{BASE-SAT}}$ performs worse than $\textsc{Tapas}_{\text{BASE}}$, which means that restricting attention to only same columns and rows by masking reduce the modeling capacity. This led to choosing soft bias addition over hard masking.

\paragraph{Attention Bias Scaling.} Unlike \textsc{TableFormer}, we also tried to add attention biases before the scaling operation in the self-attention module (SO). Specifically, we compute pair-wise attention score by:
\begin{equation}
\begin{aligned}
A_{ij} = \frac{(h_i^\top W^Q) (h_j^\top W^K)^\top + \hat{A}_{ij}}{\sqrt{d_K}}
\end{aligned}
\end{equation}
instead of using:
\begin{equation}
\begin{aligned}
A_{ij} = \frac{(h_i^\top W^Q) (h_j^\top W^K)^\top }{\sqrt{d_K}} + \hat{A}_{ij},
\end{aligned}
\end{equation}
which is the element-wise version of Equation \ref{eqp} and \ref{eqtatt}. However, Table~\ref{tab:widgets5} shows that  $\textsc{TableFormer}_{\text{BASE-SO}}$ performs worse than $\textsc{TableFormer}_{\text{BASE}}$, showing the necessity of adding attention biases after the scaling operation. 
We think the reason is that the attention bias term does not require scaling, because attention bias scalar magnitude is independent of $d_K$, while the dot products grow large in magnitude for large values of $d_K$. Thus, such bias term could play an more important role without scaling, which helps each attention head know clearly what to pay more attention to according to stronger inductive biases.

\begin{table}[t]
\small
\centering
\begin{tabular}{lcccc}
\toprule
                                          & \textit{rc-gp} & \textit{c-gp} & \textit{gp} & \textit{pcp} \\\midrule
$\textsc{Tapas}_{\text{BASE}}$                                & 57.6       & 47.4                           & 46.4                       & 29.1                        \\
$\textsc{Tapas}_{\text{BASE-SAT}}$          & 45.2       & -                              & -                          & -                           \\
$\textsc{TableFormer}_{\text{BASE-SO}}$                & 60.0       & 60.2                           & 59.8                       & 60.7                        \\
$\textsc{TableFormer}_{\text{BASE}}$                          & 62.2       & 61.5                           & 61.7                       & 61.9                        
\\\bottomrule
\end{tabular}
\caption{\label{tab:widgets5}ALL questions' cell selection accuracy of \textsc{TableFormer} variants on SQA development set. \textit{rc-gp} represents the setting including row ids, column ids and global positional ids, \textit{c-gp} represents column ids and global positional ids, \textit{gp} represents global positional ids, and \textit{pcp} represents per-cell positional ids. ``SAT'' represents masking out some attention scores. ``SO'' represents adding attention bias before scaling.}
\end{table}

\paragraph{Row, Column, \& Global Positional IDs.} With $\textsc{Tapas}_{\text{BASE}}$, $\textsc{TableFormer}_{\text{BASE-SO}}$, and $\textsc{TableFormer}_{\text{BASE}}$, we first tried the full-version where row ids, column ids, and global positional ids exist as input (\textit{rc-gp}). Then, we deleted row ids (\textit{c-gp}), and column ids (\textit{gp}) sequentially. Finally, we changed global positional ids in \textit{gp} to per-cell positional ids (\textit{pcp}). Table \ref{tab:widgets5} shows that  $\textsc{Tapas}_{\text{BASE}}$ performs significantly worse from \textit{rc-gp} $\rightarrow$ \textit{c-gp} $\rightarrow$ \textit{gp} $\rightarrow$ \textit{pcp}, because table structure information are deleted sequentially during such process. However, with $\textsc{TableFormer}_{\text{BASE}}$, there is no obvious performance drop during the same process. That shows the structural inductive biases in \textsc{TableFormer} can provide complete table structure information. Thus, row ids, column ids and global positional ids are not necessary in \textsc{TableFormer}. We pick \textsc{TableFormer} \textit{pcp} setting as our final version to conduct all other experiments in this paper. In this way, \textsc{TableFormer} is strictly invariant to row and column order perturbation by avoiding spurious biases in those original ids.

\subsection{Comparison of \textsc{TableFormer} and Perturbed Data Augmentation}

\begin{table}
\footnotesize
\centering
\begin{tabular}{lcc|cc}
\toprule
                                      & \multicolumn{2}{c|}{\textbf{Befor Perturb}}        & \multicolumn{2}{c}{\textbf{After Perturb}}       \\\midrule
                                            & ALL      & SEQ      & ALL      & $VP$             \\\midrule
$\textsc{Tapas}_{\text{BASE}}$                                     & 61.1          & 31.3          & 57.4           & 14.0\%         \\\midrule
$\textsc{Tapas}_{\text{BASE}}$ 1p                                  & 63.4          & 34.6          & 63.4                                                         & 9.9\%          \\
$\textsc{Tapas}_{\text{BASE}}$  2p                                 & 64.6          & 35.6          & 64.5                                 & 8.4\%          \\
$\textsc{Tapas}_{\text{BASE}}$  4p                                 & 65.1          & 37.0          & 65.0                                      & 8.1\%          \\
$\textsc{Tapas}_{\text{BASE}}$  8p          & 65.1          & 37.3          & 64.3                                                      & 7.2\%          \\
$\textsc{Tapas}_{\text{BASE}}$  16p                                 & 62.4          & 33.6          & 62.2                           & 7.0\%          \\\midrule
$\textsc{TableFormer}_{\text{BASE}}$                       & \textbf{66.7} & \textbf{39.7} & \textbf{66.7}        & \textbf{0.1\%}
\\\bottomrule
\end{tabular}
\caption{\label{tab:widgets6}Comparison of \textsc{TableFormer} and perturbed data augmentation on SQA test set, where $VP$ represents model prediction variation percentage after perturbation. Median of 5 independent runs are reported.}
\end{table}

As stated in Section \ref{sec:aug},  perturbing row and column orders as augmented data during training can serve as another possible solution to alleviate the spurious row/column ids bias. Table \ref{tab:widgets6} shows the performance of $\textsc{Tabpas}_{\text{BASE}}$ model trained with additional \{1, 2, 4, 8, 16\} perturbed versions of each table as augmented data. 

We can see that the performance of $\textsc{Tapas}_{\text{BASE}}$ on SQA dataset improves with such augmentation. Also, as the number of perturbed versions of each table increases, model performance first increases and then decreases, reaching the best results with 8 perturbed versions. We suspect that too many versions of the same table confuse the model about different row and column ids for the same table, leading to decreased performance from 8p to 16p. Despite its usefulness, such data perturbation is still worse than \textsc{TableFormer}, because it could not incorporate other relevant text-table structural inductive biases like \textsc{TableFormer}. 

Although, such data augmentation makes the model more robust to row and column order perturbation with smaller $VP$ compared to standard $\textsc{Tapas}_{\text{BASE}}$, there is still a significant prediction drift after perturbation. As shown in Table~\ref{tab:widgets6}, $VP$ decreases from 1p to 16p, however, the best $VP$ (7.0\%) is still much higher than (nearly) no variation (0.1\%) of \textsc{TableFormer}.

To sum up, \textsc{TableFormer} is superior to row and column order perturbation augmentation, because of its additional structural biases and strictly consistent predictions after perturbation.

\begin{table}[t!]

\centering
\small
\begin{tabularx}{\linewidth}{Xcc}
\toprule
     & ALL       & SEQ       \\ \midrule
 $\textsc{TableFormer}_{\text{BASE}}$              & \textbf{62.1} & \textbf{38.4} \\\midrule
 - Same Row              & 32.1          & 2.8           \\
 - Same Column          & \textbf{62.1} & 37.7          \\
 - Same Cell            & 61.8          & \textbf{38.4} \\
 - Cell to Column Header & 60.7          & 36.6          \\
 - Cell to Sentence          & 60.5          & 36.4          \\
 -  Header to Column Cell & 60.5          & 35.8          \\
 - Header to Other Header    & 60.6          & 35.8          \\
 - Header to Same Header     & 61.0          & 36.9          \\
 - Header to Sentence        & 61.1          & 36.3          \\
 - Sentence to Cell          & 60.8          & 36.2          \\
 - Sentence to Header        & 61.0          & 37.3          \\
 - Sentence to Sentence      & 60.0          & 35.3          \\
 -~All Column Related (\# 2, 4, 6)  &  54.5          & 29.3          \\
\bottomrule
\end{tabularx}
\caption{\label{tab:widgets7}Ablation study of proposed attention biases.}
\vspace{-1em}
\end{table}

\subsection{Attention Bias Ablation Study}

We conduct ablation study to demonstrate the utility of all 12 types of defined attention biases. For each ablation, we set the corresponding attention bias type id to \emph{``others''} bias id. Table \ref{tab:widgets7} shows $\textsc{Tapas}_{\text{BASE}}$'s performance SQA dev set. Overall, all types of attention biases help the \textsc{TableFormer} performance to some extent, due to certain performance drop after deleting each bias type. 

Amongst all the attention biases, deleting \emph{``same row''} bias leads to most significant performance drop, showing its crucial role for encoding table row structures. There is little performance drop after deleting \emph{``same column''} bias, that's because \textsc{TableFormer} could still infer the same column information through \emph{``cell to its column header''} and \emph{``header to its column cell''} biases. After deleting all same column information (\emph{``same column''}, \emph{``cell to column header''} and \emph{``header to column cell''} biases), \textsc{TableFormer} performs significantly worse without encoding column structures. Similarly, there is little performance drop after deleting \emph{``same cell''} bias, because \textsc{TableFormer} can still infer same cell information through \emph{``same row''} and \emph{``same column''} biases. 

\subsection{Limitations of \textsc{TableFormer}}

\textsc{TableFormer} increases the  training time by around $20\%$, which might not be ideal for very long tables and would require a scoped approach.
Secondly, with the strict row and column order invariant property, \textsc{TableFormer} cannot deal with questions based on absolute orders of rows in tables. This however is not a practical requirement based on the current dataset. Doing a manual study of 1800 questions in SQA dataset, we found that there are 4 questions\footnote{We find such 4 questions by manually looking at all 125 questions where the model predictions turn from correct to incorrect after replacing $\textsc{Tapas}_{\text{LARGE}}$ with $\textsc{TableFormer}_{\text{LARGE}}$.} (0.2\% percentage) whose answers depend on orders of rows. Three of them asked \emph{``which one is at the top of the table''}, another asks \emph{``which one is listed first''}. However, these questions could be potentially answered by adding back row and column order information based on \textsc{TableFormer}.

\section{Other Related Work}
\paragraph{Transformers for Tabular Data.}  
\citet{yin20acl} prepended  corresponding column headers to cells contents, and \citet{chen2019tabfact} used corresponding column headers as features for cells. However, such methods encode each table header multiple times, leading to duplicated computing overhead.  Also, tabular structures (e.g. same row information) are not fully incorporated to such models. Meanwhile, \citet{yin20acl} leveraged row encoder and column encoder sequentially, which introduced much computational overhead, thus requiring retrieving some rows as a preprocessing step. Finally, SAT \cite{zhang2020table}, \citet{deng2020turl} and \citet{wang2021tuta} restricted attention to same row or columns with attention mask, where such inductive bias is too strict that cells could not directly attend to those cells in different row and columns, hindering the modeling ability according to Table \ref{tab:widgets5}. \citet{liu2021tapex} used the seq2seq BART generation model with a standard Transformer encoder-decoder architecture.  In all models mentioned above, spurious inter-cell order biases still exist due to global positional ids of Transformer, leading to the vulnerability to row or column order perturbations, while our \textsc{TableFormer} could avoid such problem. \citet{muller2019answering} and \citet{wang2019rat} also used relative positional encoding to encode table structures, but they modeled the relations as learnable relation vectors, whose large overhead prevented pretraining and led to poor performance without pretraining, similarly to ETC \cite{ainslie2020etc} explained in Section \ref{para:ETC}. 

\paragraph{Structural and Relative Attention.} Modified attention scores has been used to model relative positions \cite{shaw2018self}, long documents \cite{dai2019transformer,Beltagy2020Longformer,ainslie2020etc}, and graphs \cite{ying2021transformers}. But adding learnable attention biases to model tabular structures has been under-explored.
\section{Conclusion}

In this paper, we identified the vulnerability of prior table encoding models along two axes: (a) capturing the structural bias, and (b) robustness to row and column perturbations. To tackle this, we propose \textsc{TableFormer}, where learnable task-independent learnable structural attention biases are introduced, while making it invariant to row/column order at the same time. Experimental results showed that \textsc{TableFormer} outperforms strong baselines in 3 table reasoning tasks, achieving state-of-the-art performance on SQA dataset, especially when facing row and column order perturbations, because of its invariance to  row and column orders.

\section*{Acknowledgments}

We thank Julian Eisenschlos, Ankur Parikh, and the anonymous reviewers for their feedbacks in improving this paper.

\section*{Ethical Considerations}

The authors foresee no ethical concerns with the research presented in this paper.





\bibliography{custom}
\bibliographystyle{acl_natbib}






\end{document}